\documentclass{article}
\usepackage{spconf,amsmath,graphicx}

\usepackage{times}
\usepackage{epsfig}
\usepackage{graphicx}
\usepackage{enumitem}
\usepackage{amssymb}

\newcommand{\metro}{$\text{MetroCon}^2\,$}

\newcommand{\sota}{\textit{state-of-the-art }}

\usepackage[pagebackref=true,breaklinks=true,colorlinks,bookmarks=false]{hyperref}

\makeatletter
\def\thickhline{%
  \noalign{\ifnum0=`}\fi\hrule \@height \thickarrayrulewidth \futurelet
   \reserved@a\@xthickhline}
\def\@xthickhline{\ifx\reserved@a\thickhline
               \vskip\doublerulesep
               \vskip-\thickarrayrulewidth
             \fi
      \ifnum0=`{\fi}}
\makeatother

\newlength{\thickarrayrulewidth}
\title{SpaceMeshLab: Spatial Context Memoization and Meshgrid Atrous Convolution Consensus for Semantic Segmentation}
\name{Taehun Kim \qquad Jinseong Kim \qquad Daijin Kim}
\address{Dept. of Computer Science and Engineering, Pohang University of Science and Technology, Korea}
\begin{document}
%
\maketitle
\begin{abstract}
Semantic segmentation networks adopt transfer learning from image classification networks causing a shortage of spatial context information. For this reason, we propose Spatial Context Memoization (SpaM), a bypassing branch for spatial context by retaining the input dimension and constantly communicating its spatial context and rich semantic information mutually with the backbone network. Multi-scale context information for semantic segmentation is crucial for dealing with diverse sizes and shapes of target objects in the given scene. Conventional multi-scale context scheme adopts multiple effective receptive fields by multiple dilation rates or pooling operations, but often suffer from misalignment problem with respect to the target pixel. To this end, we propose Meshgrid Atrous Convolution Consensus (\metro) which brings multi-scale scheme into fine-grained multi-scale object context using convolutions with meshgrid-like scattered dilation rates. SpaceMeshLab (ResNet-101 + SpaM + \metro) achieves 82.0\% mIoU in Cityscapes $test$ and 53.5\% mIoU on Pascal-Context $val$ set.
\end{abstract}
\section{Introduction}
\label{sec:intro}

Modern computer vision tasks utilize powerful semantic features from image classification networks with pre-trained parameters on ImageNet \cite{russakovsky2015imagenet}. Such networks reduce a spatial dimension and increase the depth of feature maps by applying striding in convolution layers or pooling operation. One of the main reasons for dimensional reduction is to ensure position and rotation invariance and reduce the number of parameters while increasing depth to balance out the parameter-depth trade-off. However, semantic segmentation requires an output with fully retained spatial dimension of an input image. To this end, we propose a detour of backbone feature maps with maintaining the original spatial dimension of the input image called \textbf{Spa}tial context \textbf{M}emoization (SpaM). We branch out feature maps from the backbone and use pixel shuffle \cite{shi2016real} and its inverse operation, pixel un-shuffle to aggregate both original feature maps from the backbone and branched out feature maps from SpaM mutually across each blocks in the backbone network.

Another problem in semantic segmentation is a misalignment of multi-scale context modules from target objects. Atrous spatial pyramid pooling (ASPP) from DeepLabV3+ \cite{chen2018encoder} uses multiple atrous convolution layers \cite{chen2017deeplab} with different dilation rates and pyramid pooling module (PPM) from PSPNet \cite{zhao2017pyramid} uses multiple grid poolings with different grid sizes in a parallel manner in order to extract multi-scale context effectively. While large dilation rate in ASPP can consider larger region than regular convolution, it causes a huge gap between each target pixel and its neighboring pixels. Therefore, we propose \textbf{Me}shgrid A\textbf{tro}us \textbf{Con}volution \textbf{Con}sensus (\metro) module to consider a fine-grained multi-scale object context using meshgrid scheme for setting dilation rates of convolution and train the contribution score for each convolution filter with different dilation rates to effectively extract multi-scale features and their consensus information.

SpaceMeshLab (ResNet-101 + SpaM + \metro) achieves 82.0\% mean Intersection over Union (mIoU) on Cityscapes \cite{cordts2016cityscapes} $test$ and achieve 53.5\% mIoU on Pascal-Context \cite{mottaghi2014role} $val$ set.

\begin{figure*}
\begin{centering}
   \includegraphics[width=0.9\linewidth]{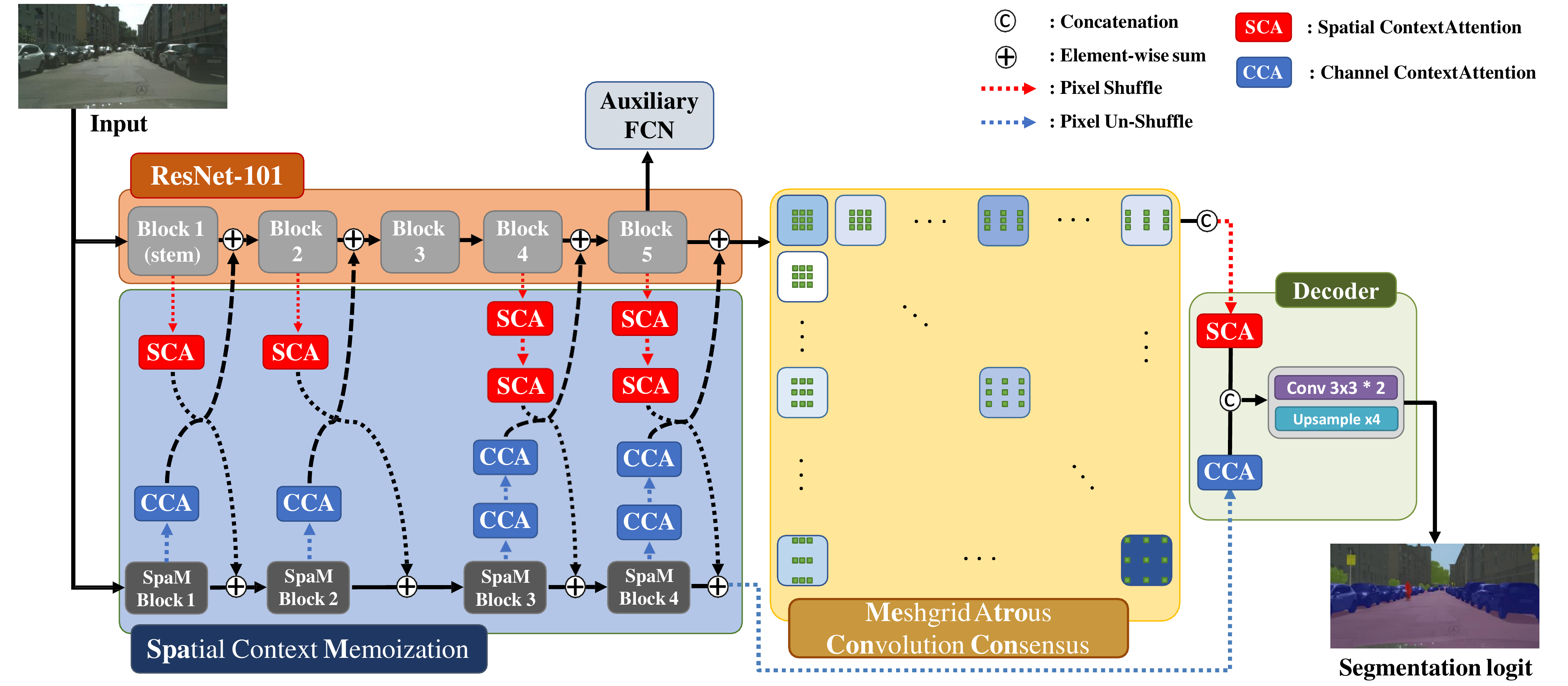}
   \caption{The overview of proposed SpaceMeshLab with ResNet-101 as a backbone network.}
\end{centering}
\label{fig:2}
\vspace{-0.5em}
\end{figure*}

\section{Method}
\label{sec:method}

\subsection{Spatial Context Memoization}
\label{spam}

SpaM consists of SpaM blocks, pixel shuffle/un-shuffle operations and self-attention modules. To preserve the spatial context information dramatically while minimizing the increase of the number of operation and memory consumption, SpaM branches out a same number of SpaM blocks, which consists of consecutive $3\times3$ convolution, batch normalization and activation layer, as a backbone network blocks to share the dense semantic context information from the backbone network and spatial contextual information from itself mutually (see \figureautorefname~\ref{fig:2}). To share features from backbone to SpaM blocks, we adopt pixel shuffle followed by Spatial Context Attention (SCA) to rearrange backbone features to the spatial size of SpaM feature. Likewise, we adopt pixel un-shuffle followed by Channel Context Attention (CCA) to rearrange SpaM features to the spatial size of backbone features. We compose two consecutive pixel shuffle (un-shuffle) and SCA (CCA) modules for gradual increment (decrement) and computational efficiency.

\subsection{Meshgrid Atrous Convolution Consensus}
\label{metro}

\metro extends multi-scale scheme with meshgrid-like dilation strategy. Atrous convolution can change the effective receptive field without increasing computations and memory consumption. However, large dilation rate can introduce irrelevant pixels with respect to the center pixel especially for narrow objects like pole. To fully cover the field of view while correctly assigning each scale context more effectively, we divide dilation rate into every possible combinations in a range of desired receptive field. We denote the dilation rates as $\textbf{d} = \{(i,j)|i\in\{i_1, ... i_M\}, j\in\{j_1, ... j_N\}\}$ where $i$ and $j$ are dilation rate for vertical and horizontal axis respectively. We empirically set dilation rates as $\textbf{d} = \{(i,j)|i, j\in\{1, 2, ... 18\}\}$ by conducting ablation study on different settings of \metro in \sectionautorefname~\ref{ablation_metro}.

Moreover, if we set a lot of different dilation rates, we cannot distinguish between the most useful dilation rate and the least because each convolution operation with different dilation rates is in charge of the same amount of data in the representation. To determine which feature map computed from certain dilation rate is more important, we add a trainable parameter for each convolution within \metro as a confidence score. Also, we adjust the number of output channels for each convolution by $\lfloor \frac{1280}{M \times N} \rceil$, where $M$ and $N$ are the number of horizontal and vertical dilation rates and $\lfloor \cdot \rceil$ denotes a round operator, to make the number of parameters and computation almost same as ASPP. Finally, the output of each convolution in \metro is concatenated and forwarded to the decoder.

\subsection{Spatial and Channel Context Attention}
\label{attn}

\begin{figure}[t]
    \begin{center}
       \includegraphics[width=0.9\linewidth]{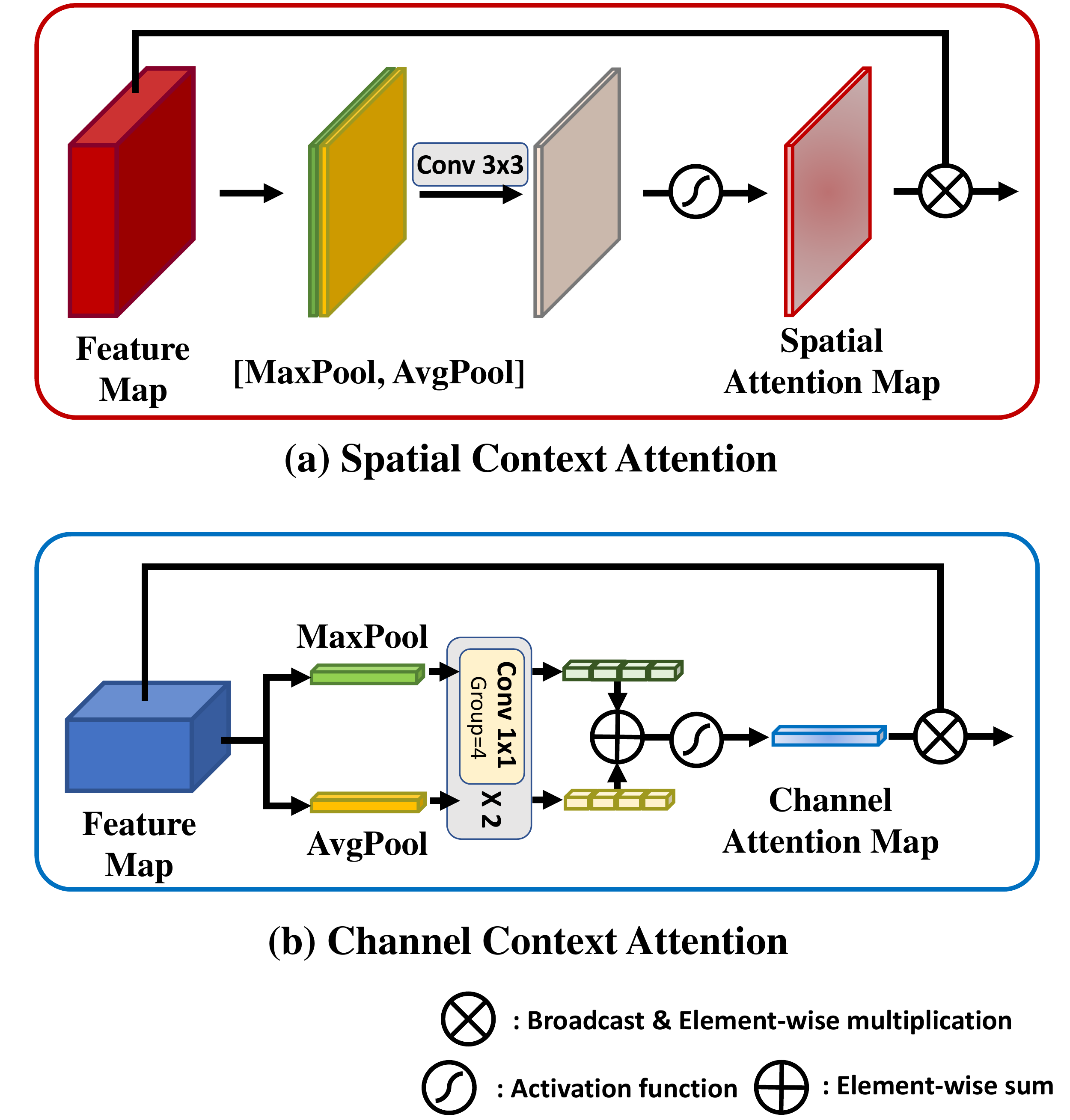}
    \end{center}
       \caption{A structure of Spatial Context Attention (SCA) and Channel Context Attention (CCA) from SpaM.} 
    \label{fig:attn}
    \vspace{-0.5em}
\end{figure}

We design SCA and CCA inspired by CBAM \cite{woo2018cbam} (see \figureautorefname~\ref{fig:attn}). While CBAM performed spatial and channel attention consecutively, we separate two different modules and apply SCA (CCA) after the pixel shuffle (un-shuffle) since explicit pixel mapping to the spatial dimension and channel dimension like pixel shuffle and un-shuffle can cause checkerboard artifact and feature disarrangement respectively. Since we do not require self-attention for every layer of the network, we modify SCA and CCA by using groups in convolutions as a multi-head manner to attend to different aspects from multiple feature groups. We conduct experiments on SpaM with and without attention modules in \sectionautorefname~\ref{ablation_spam}.

\subsection{SpaceMeshLab Decoder}
\label{decoder}
While we try to extract spatial context and multi-scale context from SpaM and \metro, we need a suitable decoder design to fully utilize their representations. The shape of output feature map from \metro is fairly conventional since the spatial dimension is same as the backbone feature, but the shape of output feature map from SpaM has the same spatial dimension of the input image. To incorporate two features, we bring pixel shuffle (un-shuffle) and SCA (CCA) modules from SpaM to rearrange two features. We then forward concatenated two rearranged features to the consecutive two $3 \times 3$ convolution layer and bilinear upsampling.

\section{Experiment}
\label{experiment}

We evaluate our proposed methods on Cityscapes \cite{cordts2016cityscapes} and Pascal-Context \cite{mottaghi2014role}. Cityscapes is an urban street scene dataset collected from 50 different cities. There are 2,375 images for training, 500 images for validation and 1,525 images for test, which are annotated for 19 classes including car, building and pole. There are also coarse annotation for extra 20,000 images but we do not use for training our models. Pascal-Context is finely annotated for the whole scene on VOC2010 dataset images. There are 4,998 images for training and 5,105 images for test. They annotated for more than 400 classes, but we follow \cite{zhang2018context} by using frequent 59 classes and ignoring other classes.

\begin{table}[]
    \begin{center}
    \resizebox{0.95\linewidth}{!}{\begin{tabular}{l|ccccc}
        \thickhline
        \textbf{Backbone} & \textbf{SpaM} & \textbf{SCA} & \textbf{CCA} & MS + Flip  & \textbf{mIoU}(\%) \\ \hline \hline
        ResNet-50         &               &              &              &            & 77.4              \\
        ResNet-50         & \checkmark    &              &              &            & 78.2              \\
        ResNet-50         & \checkmark    & \checkmark   & \checkmark   &            & 78.7              \\ \hline
        ResNet-101        &               &              &              &            & 78.1              \\
        ResNet-101        &               &              &              & \checkmark & 79.5              \\
        ResNet-101        & \checkmark    &              &              &            & 78.9              \\
        ResNet-101        & \checkmark    & \checkmark   & \checkmark   &            & 79.4              \\
        ResNet-101        & \checkmark    & \checkmark   & \checkmark   & \checkmark & \textbf{80.5}     \\ \thickhline
    \end{tabular}}
    \end{center}
    \vspace{-0.5em}
    \caption{Ablation study of SpaM on Cityscapes \textit{val} set. Note that we denote \textbf{SpaM}  without \textbf{SCA} and \textbf{CCA} if both module are not selected as \checkmark and ``MS + Flip'' denote testing time augmentation.}
    \label{tab:1}
    
\end{table}

\begin{table}[]
    \begin{center}
    \resizebox{\linewidth}{!}{\begin{tabular}{l|cccc}
        \thickhline
        \textbf{Method}   &  Dilation Rates               & MS + Flip  & Depth   & \textbf{mIoU}(\%) \\ \hline \hline
        ASPP              &  $\cdot$                      &            & $\cdot$ & 79.4              \\
        ASPP              &  $\cdot$                      & \checkmark & $\cdot$ & 80.5              \\
        \metro            & $i, j\in\{6, 12, 18\}$        &            &  144    & 79.8              \\
        \metro            & $i, j\in\{1, 2, \dots, 9\}$   &            &  16     & 80.4              \\ 
        \metro            & $i, j\in\{1, 2, \dots, 18\}$  &            &  4      & 80.8     \\
        \metro            & $i, j\in\{1, 2, \dots, 18\}$  & \checkmark &  4      & \textbf{81.8}     \\\thickhline
    \end{tabular}}
    \end{center}
    \vspace{-0.5em}
    \caption{Ablation study of \metro on Cityscapes \textit{val} set. Both ASPP and \metro are deployed on ResNet-101 with SpaM.}
    \label{tab:2}
    \vspace{-0.5em}
\end{table}

We implement our methods with Pytorch \cite{paszke2019pytorch}. As shown in \figureautorefname~\ref{fig:2}, we design SpaM on top of ResNet-50 and ResNet-101 with atrous convolution strategy \cite{chen2017deeplab, zhao2017pyramid}. We set a scaling factor of pixel shuffle (un-shuffle) in SpaceMeshLab as 4 and we omit ``Block 3'' corresponding stage in SpaM for consistent pixel shuffle (un-shuffle) operation. We follow the basic training strategy from \cite{chen2017rethinking}. We set dilation rates in \metro as $\textbf{d} = \{(i,j)|i, j\in\{1, 2, ... 18\}\}$ if not specified. We add auxiliary FCN branch on the last block of the backbone which has one additional convolution layer for auxiliary loss. We use standard cross entropy loss for both final loss and auxiliary loss. The weight for auxiliary loss is 0.4. For testing time augmentation, we resize image with \{0.5, 0.75, 1.0, 1.25, 1.5, 1.75\} and flip horizontally for each scale. To obtain semantic logit, we average results from each sample by resizing to the original scale and flip again if flipped. We denote testing time augmentation as ``MS + flip''. We set the initial learning rate as 0.01, 80K maximum iterations and a batch size of 8.

\subsection{Ablation Study for Spatial Context Memoization}
\label{ablation_spam}
We conduct experiments on SpaM with different settings as shown in \tableautorefname~\ref{tab:1} to verify the effect of SpaM, SCA and CCA on Cityscapes $val$ set. We first add SpaM on top of the baseline (DeepLabV3+) without using attention modules, SCA and CCA. We then add attention modules after pixel shuffle and un-shuffle. For ResNet-50 backbone, SpaM without SCA and CCA show 0.8\% mIoU improvement and adding SCA and CCA modules improve extra 0.5\% mIoU. For deeper backbone model ResNet-101, SpaM show better performance by 0.8\% mIoU, and adding SCA and CCA show 79.4\% mIoU which is almost same performance as the baseline with testing time augmentation. With testing time augmentation, SpaM with SCA and CCA show 80.5\% mIoU.

\subsection{Ablation Study for Meshgrid Atrous Convolution Consensus}
\label{ablation_metro}
We analyze different settings of \metro by changing the numbers of dilation rates on Cityscapes $val$ set. In \tableautorefname~\ref{tab:2}, we bring same numbers of dilation rates from \cite{chen2018encoder} which are 6, 12 and 18. With 3 different numbers for both horizontal and vertical dilation, we have 9 different combinations for \metro, so we change the number of output channels for each convolution in \metro as 144 to match the total output channels as close as the number of output channels from ASPP. We denote the number of output channels for each convolution in \metro as ``Depth'' in \tableautorefname~\ref{tab:2}. We achieve 79.8\% mIoU which is 0.4\% improvement compared to the ASPP module. We further explore to find better dilation rates combinations by changing the numbers for \metro. Using \{1, 2, ..., 9\} and \{1, 2, ..., 18\} for both horizontal and vertical dilation rates, We can obtain 81 and 324 different combinations respectively and we set the number of output channels for each convolution in \metro as 16 and 4 respectively which makes the final the number of output channels equals to the  previous setting. We achieve 80.4\% mIoU and 80.8\% mIoU for each setting and with testing time augmentation for \{1, 2, ..., 18\} setting, we achieve 81.8\% mIoU which improve 2.3\% mIoU compared to the DeepLabV3+. Since \{1, 2, ..., 18\} shows best performance, we use this setting for our default setting of SpaceMeshLab.

\begin{table}[]
    \begin{center}
    \resizebox{0.75\linewidth}{!}{\begin{tabular}{l|cc}
        \thickhline
        \textbf{Method} & MS + Flip &  \textbf{mIoU}(\%)            \\ \hline \hline
        FCN-8s \cite{long2015fully}       & \checkmark & 65.3 \\
        DeepLabv2 \cite{chen2017deeplab}  & \checkmark & 70.4 \\
        RefineNet \cite{lin2017refinenet} & \checkmark & 73.6 \\
        DUC \cite{wang2018understanding}  & \checkmark & 77.6 \\
        PSPNet \cite{zhao2017pyramid}     & \checkmark & 78.4 \\
        DenseASPP \cite{yang2018denseaspp}& \checkmark & 80.6 \\
        DANet \cite{fu2019dual}           & \checkmark & 81.5 \\
        SpyGR \cite{li2020spatial}        & \checkmark & 81.6 \\ \hline \hline
        SpaceMeshLab ($Ours$)         &      & 80.9     \\
        SpaceMeshLab ($Ours$)         & \checkmark     & \textbf{82.0}     \\
        
        \thickhline
    \end{tabular}}
    \end{center}
    \vspace{-0.5em}
    \caption{Comparison of SpaceMeshLab with \sota methods on Cityscapes \textit{test} set.}
    \label{tab:3}
    
\end{table}

\begin{table}[]
    \begin{center}
    \resizebox{0.75\linewidth}{!}{\begin{tabular}{l|cc}
        \thickhline
        \textbf{Method}      & MS + Flip & \textbf{mIoU}(\%) \\ \hline \hline
        FCN \cite{long2015fully} & \checkmark& 37.8 \\
        DeepLabv2 \cite{chen2017deeplab} &\checkmark & 45.7 \\
        PSPNet \cite{zhao2017pyramid} & \checkmark& 47.8 \\
        $\text{DeepLabV3+}^\dagger$ \cite{chen2018encoder} & \checkmark & 51.7 \\
        DANet \cite{fu2019dual} & \checkmark & 52.6 \\
        SpyGR \cite{li2020spatial} & \checkmark & 52.8 \\\hline \hline
        $\text{SpaceMeshLab}^\dagger$($Ours$) &            & 52.8    \\ 
        $\text{SpaceMeshLab}^\dagger$($Ours$) & \checkmark & \textbf{53.5}     \\ \thickhline
    \end{tabular}}
    \end{center}
    \vspace{-0.5em}
    \caption{Comparison of SpaceMeshLab with \sota methods on on Pascal-Context $val$ set. $\dagger$ denotes our implementation.}
    \label{tab:4}
    \vspace{-0.8em}
\end{table}

\subsection{Qualitative Results and Evaluation with State-of-the-art Methods}
\begin{figure}[t]
    \begin{center}
       \includegraphics[width=1.0\linewidth]{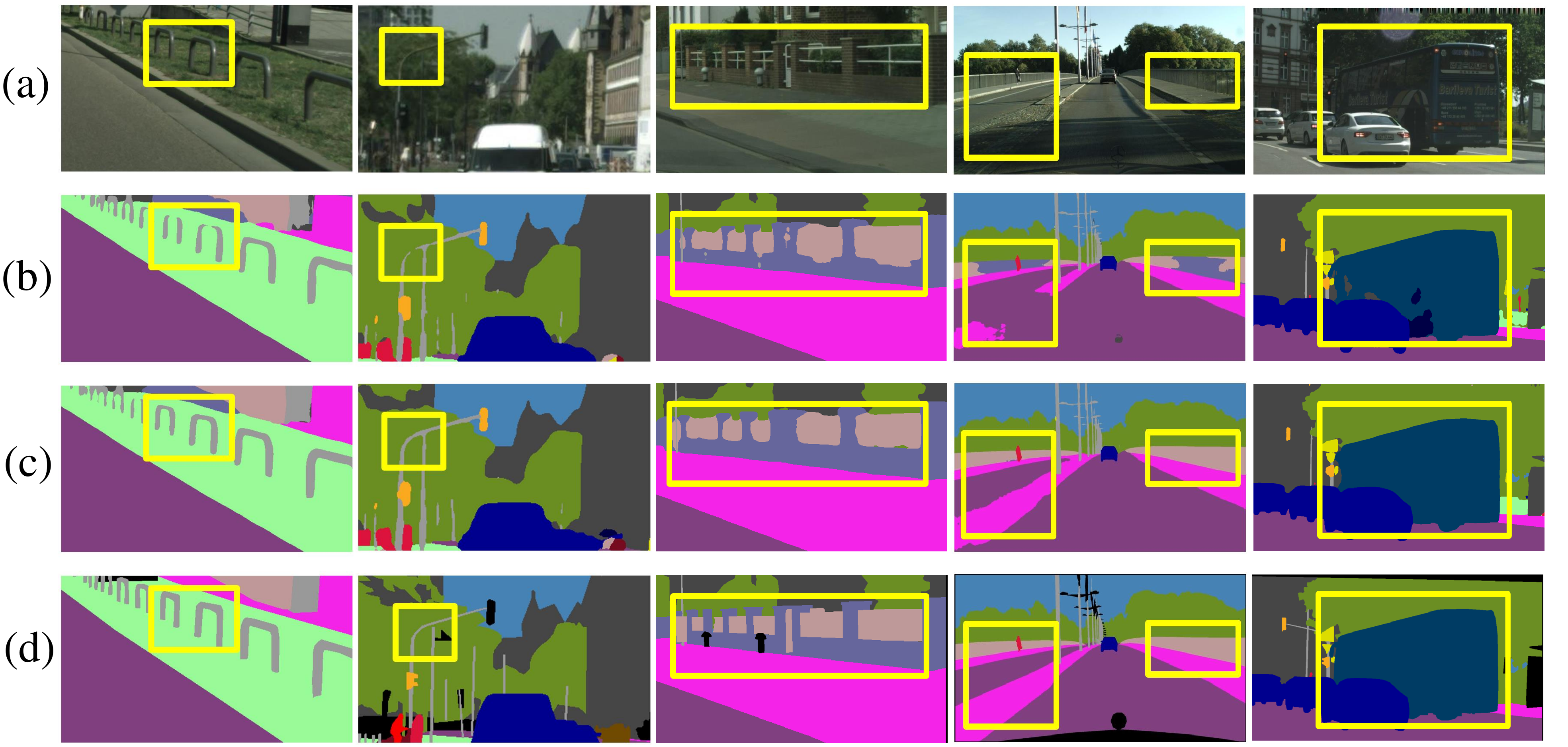}
    \end{center}
       \caption{Qualitative results of baseline (DeepLabV3+) and SpaceMeshLab on Cityscapes dataset. (a): input image. (b): segmentation results of baseline. (c): segmentation results of SpaceMeshLab. (d): ground truth.} 
    \label{fig:5}
    \vspace{-0.5em}
\end{figure}

In the first and second column in \figureautorefname~\ref{fig:5}, the baseline shows some disconnected parts on sidewalk poles and traffic light pole while SpaceMeshLab successfully segments proper regions without any disconnection. This example shows that SpaceMeshLab is more effective for capturing multi-scale context by considering variety of regions compared to ASPP. Also, in the third and fourth columns in \figureautorefname~\ref{fig:5}, our model shows better quality of discriminating wall (grey) and fence (apricot) than baseline network. This example shows that SpaceMeshLab is better for preserving the spatial context.

As shown in \tableautorefname~\ref{tab:3}, SpaceMeshLab achieves 82.0\% mIoU on Cityscapes $test$ set which show better performance over previous \textit{state-of-the-art} models especially for considering classes which occupy only a small portion in the dataset such as ``fence'' or ``pole''. We also conduct another experiment on Pascal-Context $val$ set to compare with our baseline network (\tableautorefname~\ref{tab:4}). While Cityscapes dataset has an identical input size, Pascal-Context varies in size, so resize an input size to $(\lfloor \frac{H}{16} \rfloor \times 16, \lfloor \frac{W}{16} \rfloor \times 16)$ where $(H,W)$ is an original size. We achieve 53.5\% mIoU which is 1.6\% improvement compared to the baseline applying testing time augmentation.

\section{conclusion}
\label{conclusion}

In this paper, we solve spatial dimension reduction problem and pixel misalignment problem from the multi-scale context scheme methods. SpaM receives rich semantic information from the backbone network and deliver spatial context to the backbone network by retaining the spatial dimension. \metro deploys every possible combination of dilation rates to each convolution to cover the entire neighboring pixels within the receptive field and assign confidence score for each convolution to gather the consensus information among them. We demonstrate the effectiveness of SpaceMeshLab both quantitatively and qualitatively on Cityscapes and Pascal-Context dataset.

\section{acknowledgement}
This work was supported by Institute of Information \& communications Technology Planning \& Evaluation (IITP) grant funded by the Korea government(MSIT) (No.2017-0-00897, Development of Object Detection and Recognition for Intelligent Vehicles) and (No.2018-0-01290, Development of an Open Dataset and Cognitive Processing Technology for the Recognition of Features Derived From Unstructured Human Motions Used in Self-driving Cars)

\bibliographystyle{IEEEbib}
\bibliography{egbib}

\end{document}